\documentclass[letterpaper, 10 pt, conference]{ieeeconf}  
\IEEEoverridecommandlockouts  

\usepackage{microtype}
\usepackage{amsmath,amssymb,amsfonts}
\usepackage{algorithmic}
\usepackage{graphicx}
\usepackage{textcomp}
\usepackage{xcolor}
\usepackage{url}
\usepackage[style=numeric, sorting=none, backref=true]{biblatex}
\DefineBibliographyStrings{english}{
  backrefpage={},
  backrefpages={}
}

\usepackage{xpatch}
\DeclareFieldFormat{backrefparens}{\addperiod{#1}}
\xpatchbibmacro{pageref}{parens}{backrefparens}{}{}
\usepackage{hyperref}
\usepackage{booktabs}
\usepackage{multirow}
\usepackage{cuted}
\usepackage{caption}
\hypersetup{
    colorlinks=true,
    linkcolor=red,
    citecolor=blue,
    filecolor=black,
    urlcolor=black,
    breaklinks=true,
    pdfborder={0 0 0}
}
\usepackage{arydshln}
\usepackage{color, colortbl}
\usepackage{float}
\def\BibTeX{{\rm B\kern-.05em{\sc i\kern-.025em b}\kern-.08em
    T\kern-.1667em\lower.7ex\hbox{E}\kern-.125emX}}

\definecolor{LightCyan}{rgb}{0.88,1,1}

\begin{document}

\title{\LARGE \bf Observer–Actor: Active Vision Imitation Learning\\ with Sparse-View Gaussian Splatting}

\author{Yilong Wang, Cheng Qian, Ruomeng Fan, and Edward Johns}
\maketitle

\begin{strip}
\centering
\includegraphics[width=1.\textwidth]{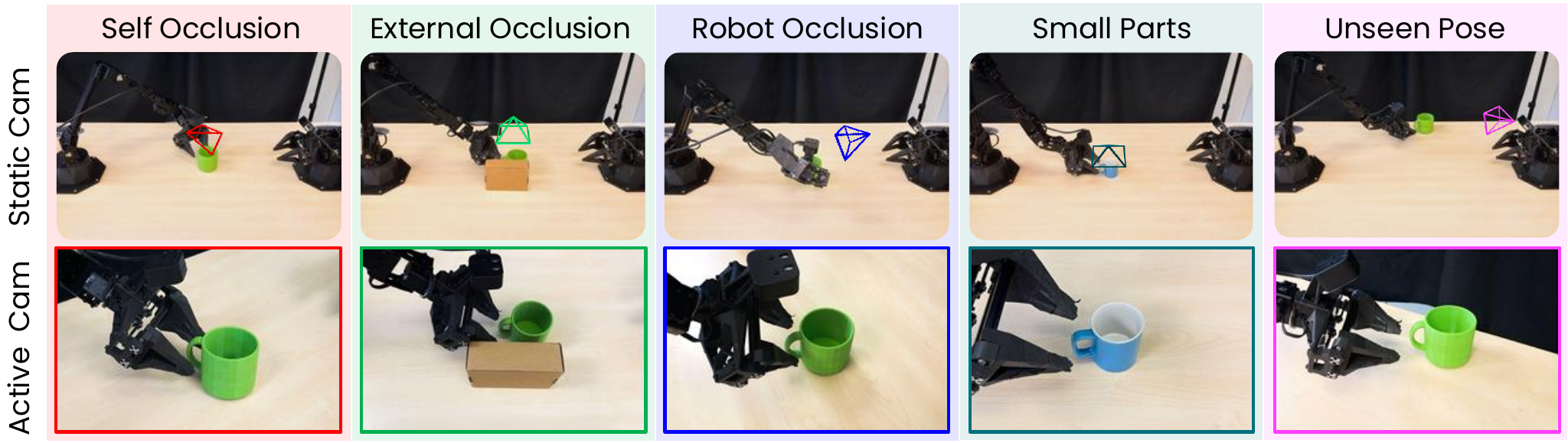}
\captionof{figure}{\textbf{Active vision for imitation learning in a mug-handle pickup task across five scenarios.} When a static camera struggles (top row), alternative placements (bottom row, and coloured frustums in top row) provide better observations. In our method, at test time an \emph{observer} robot (robot on right in above examples) computes and moves to such an optimal view from its wrist-cam, after which an \emph{actor} robot (robot on left in above examples) performs the task conditioned on this view.}

\vspace{-10pt}
\label{fig:intro}
\end{strip}

\begin{abstract}
We propose \textbf{Observer-Actor} (ObAct), a novel framework for active vision imitation learning in which the observer moves to optimal visual observations for the actor. We study ObAct on a dual-arm robotic system equipped with wrist-mounted cameras. At test time, ObAct dynamically assigns \emph{observer} and \emph{actor} roles: the observer arm constructs a 3D Gaussian Splatting (3DGS) representation from three images, virtually explores this to find an optimal camera pose, then moves to this pose; the actor arm then executes a policy using the observer’s observations. This formulation enhances the clarity and visibility of both the object and the gripper in the policy's observations. As a result, we enable the training of ambidextrous policies on observations that remain closer to the occlusion-free training distribution, leading to more robust policies. We study this formulation with two existing imitation learning methods -- trajectory transfer and behaviour cloning -- and experiments show that ObAct significantly outperforms static-camera setups: trajectory transfer improves by 145\% without occlusion and 233\% with occlusion, while behavior cloning improves by 75\% and 143\%, respectively. Videos are available at 
\begingroup \hypersetup{urlcolor=blue} \url{https://obact.github.io}. 
\endgroup
\end{abstract}

\section{Introduction}
Current imitation learning methods for robotic manipulation primarily rely on static cameras \cite{chi2023diffusion, doi:10.1126/scirobotics.adv7594}, egocentric wrist-mounted cameras \cite{chi2024universal, di2024dinobot}, or a combination of both \cite{zhao2023learning, wang2025one}. Static cameras are typically placed in a task-agnostic manner, and once a policy is trained, it often must be deployed in the same setup. Wrist-mounted cameras offer greater flexibility but suffer from limited global awareness and a restricted field of view. Combining both viewpoints can partially mitigate these issues but often introduces redundant observations that may distract the model—especially in low-data regimes \cite{chuang2024active, xiong2025vision}. These limitations are further amplified by occlusions, highlighting the need for active vision, where the camera can move dynamically to provide the best viewpoint for the task.

Realizing active vision requires both movable camera hardware and sophisticated active vision strategies. Recent studies \cite{chuang2024active, xiong2025vision, pmlr-v270-cheng25b} achieve this by developing a dedicated active-vision arm for perception, with its policy learned through teleoperation and behavior cloning. However, this approach has two significant limitations. First, the active vision arm is restricted to a fixed observer role, limiting its reachable viewpoints and preventing its use as regular manipulator. Second, the active vision strategy demands extensive human demonstrations, which, although collectable via a VR-based teleoperation system, imposes additional burden on operators.

In this paper, we present ObAct (short for Observer–Actor), a novel framework for active vision in imitation learning, where an observer robot computes optimal visual observations that guide the actor robot’s actions. As illustrated in Figure~\ref{fig:intro} with a mug-handle pickup task, ObAct positions the observer arm at test-time optimal viewpoints, providing visual observations for the actor arm during manipulation. Unlike prior approaches, ObAct dynamically assigns observer and actor roles based on the scene configuration, without requiring a separately trained active vision strategy for each arm. To enable this, we construct a test-time 3D Gaussian Splatting (3DGS) representation~\cite{kerbl20233d} from sparse-view images and optimize the camera pose to obtain observations that are similar to the demonstration observations, while minimizing occlusions. We further extend existing imitation learning methods—trajectory transfer \cite{doi:10.1126/scirobotics.adv7594} and behavior cloning \cite{zhao2023learning}—to leverage this view-conditioned framework. Experiments show that ObAct substantially improves success rates over static-camera baselines: trajectory transfer increases by 145\% without occlusion and 233\% with occlusion, while behavior cloning increases by 75\% and 143\%, respectively.

In summary, our contributions are threefold:
\begin{enumerate}
    \item \textbf{ObAct Framework}: We introduce a decoupled observer–actor framework for active vision imitation learning, which allows the system to be robust against visual edge cases that static cameras cannot.
    \item \textbf{Active Vision via Sparse-view 3DGS}: We develop an RGB-based active vision system that uses a test-time 3DGS model, constructed from sparse-view images, to optimize viewpoints for demonstration consistency and occlusion reduction. To our knowledge, this is the first use of sparse-view 3DGS in active vision.
    \item \textbf{Active Vision for Imitation Learning}: We extend trajectory transfer and behavior cloning to the active vision setting, demonstrating substantial performance gains and, for behavior cloning, improved data efficiency.
\end{enumerate}

\begin{figure*}[t!]
\centering
\includegraphics[width=.95\textwidth]{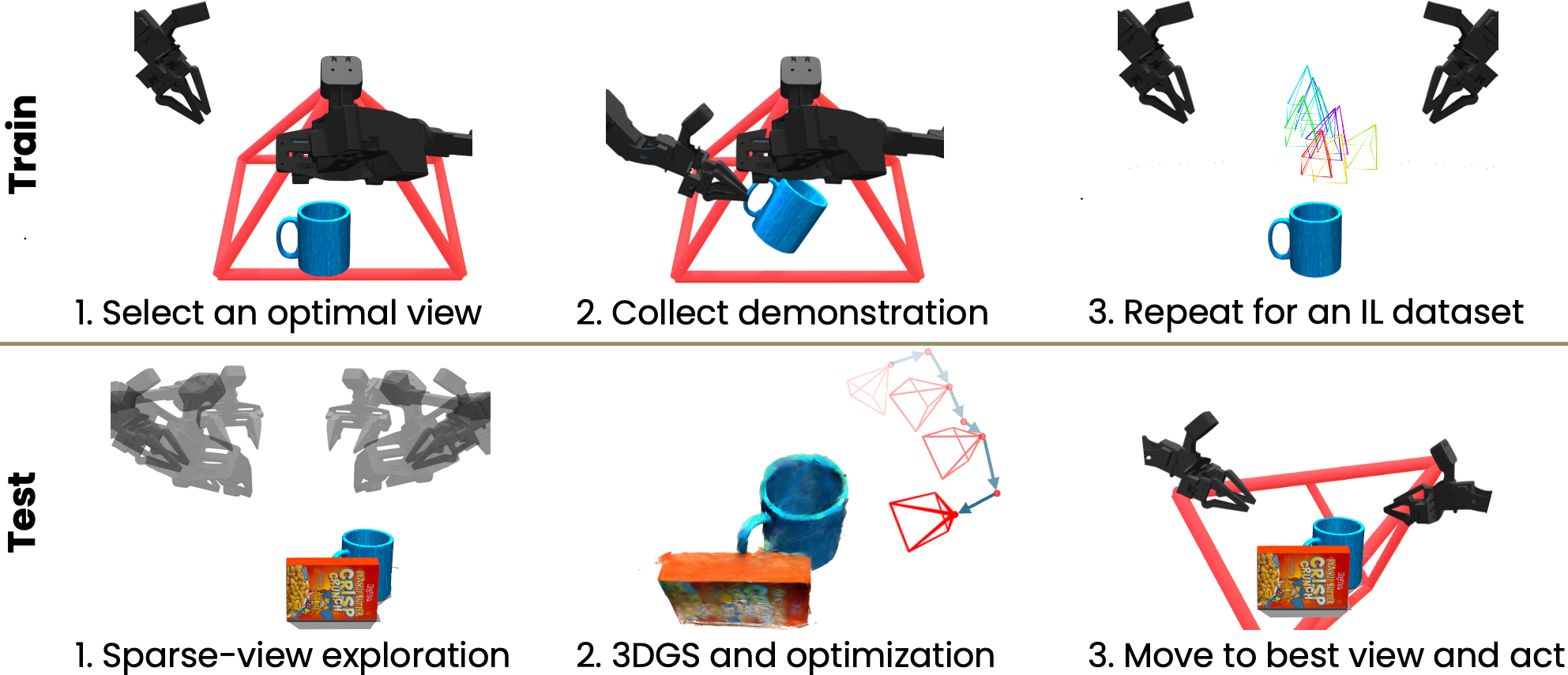}
\caption{
\textbf{Framework Overview.} 
\textbf{(1) Train:} The operator selects a \emph{demonstration optimal view}, moves the observer arm to this view, and records a demonstration. This process is repeated as required by the imitation learning method.
\textbf{(2) Test:} The robots explore six views of the scene to construct a 3DGS representation. View optimization within this representation identifies the \emph{test-time optimal view}. The observer arm then moves to this view, after which the actor arm executes the task.}
\label{overall}
\vspace{-10pt}
\end{figure*}

\section{Related Work}
\textbf{Active Vision for Robotic Manipulation.} 
Active vision (AV) \cite{aloimonos1988active} refers to actively moving the camera to maximize perceptual quality for downstream tasks, and has been widely applied in robotics for object recognition \cite{7139782}, pose estimation \cite{7139782, an2024rgbmanip}, and 3D reconstruction \cite{7487527, lee2022uncertainty}. It has also been extensively studied in robotic grasping, where unobserved views are sampled and evaluated using depth images based on information gain \cite{breyer2022closed}, grasp affordances \cite{pmlr-v229-zhang23i}, graspness inconsistency \cite{NEURIPS2024_4364fef0}, and velocity fields \cite{shi2025viso}, yielding improved performance in occluded and cluttered environments. In contrast, our method renders and evaluates unobserved viewpoints directly in the RGB domain, the prevalent modality in modern imitation learning systems \cite{chi2023diffusion, zhao2023learning}, and extends beyond grasping to support a broader range of manipulation tasks. Our setup is most closely related to \cite{wang2025observe}, where an agent first observes the environment and then interacts with it by jointly training a next-best-view camera policy and a next-best-pose gripper policy using few-shot reinforcement learning. However, their approach assumes a freely moving camera and is validated only in simulation, whereas we demonstrate a real-world dual-arm system that achieves active vision without requiring a separate policy. In concurrent work, \cite{yang2025mobipi} also avoid training by leveraging 3DGS to build a scene model from dense scans and applying Bayesian optimization to determine an initial robot pose for task execution on a mobile platform. However, our focus is on sparse-view inputs in a dual-arm setting. \cite{kerrj2025eyerobot} capture demonstrations with a 360° camera and jointly train an eyeball gaze controller and arm policies using a BC–RL loop, but their 2-DoF gaze mechanism is less capable of handling occlusions compared to our 6-DoF observer arm.

\textbf{View-conditioned Imitation Learning.} 
For manipulation systems to generalize, visuomotor policies must remain robust to discrepancies in observational viewpoints—particularly unseen camera poses—which is especially critical in active vision settings where camera positions are dynamic rather than static. Recent works address this by training end-to-end behavior cloning policies that jointly control both the active camera and the manipulation arms using human demonstrations \cite{xiong2025vision, pmlr-v270-cheng25b, chuang2024active}. While effective, these approaches increase data requirements and model complexity. In contrast, our approach trains policies over a compact distribution of viewpoints and explicitly optimizes the test-time viewpoint relative to demonstrations. Another class of methods builds explicit 3D scene representations, such as point clouds or voxels \cite{shridhar2023perceiver, ze20243d, gervet2023act3d}, or renders canonical 2D views from these structures \cite{goyal2023rvt}. These approaches improve robustness to viewpoint variation but require precise extrinsic calibration and incur high computational costs. Data augmentation offers a complementary direction: \cite{tian2024view} generate multi-view datasets from single-view demonstrations using a pre-trained zero-shot novel-view synthesis model \cite{sargent2023zeronvs}, while \cite{yang2025novel, ren2025learning} edit reconstructed 3D scenes built from dense scans to create synthetic training views. Beyond end-to-end training, some works transfer a single demonstration trajectory at test time by estimating relative object pose changes between demonstration and execution views, either via pose estimation \cite{doi:10.1126/scirobotics.adv7594} or visual servoing \cite{wang2025one}. However, such methods remain highly sensitive to occlusions. In this paper, we extend behavior cloning and trajectory transfer within our observer–actor framework, demonstrating consistent performance gains in both non-occluded and occluded scenarios.

\section{Method}

Figure~\ref{overall} illustrates the proposed system framework. To collect a demonstration, the operator selects a demonstration optimal viewpoint (Section~\ref{optimal view}) for the task object, positions the observer accordingly, and records a trajectory using the actor arm, creating a dataset for downstream imitation learning. At test time, given a novel object configuration with potential occlusions, each arm captures three predefined scene views, yielding six views in total. The system dynamically assigns observer and actor roles based on these views and constructs a 3DGS representation from the observer’s three views (Section~\ref{sparse gaussian}). Through view optimization (Section~\ref{view optimization}), it identifies the test-time optimal viewpoint within this representation. The observer arm then moves to this viewpoint, allowing the actor arm to execute the manipulation task conditioned on it (Section~\ref{imitation learning}).

\subsection{Definition of Optimal View}
\label{optimal view}

\noindent \textbf{Demonstration Optimal View.}  
We define a demonstration optimal view as a camera viewpoint relative to the target object that maximizes the visibility of task-relevant features while minimizing occlusion. The actor’s interaction with the object must remain clearly observable from this view, without obstruction from the actor arm during manipulation. In practice, demonstration optimal views avoid self-occlusion, robot-induced occlusion while preserving fine object details, as illustrated in Figure~\ref{fig:intro}. These views are determined by the operator using their own subjective estimation of where an optimal view is. We denote the demonstration optimal view as
\(
v^*_{\mathrm{demo}} \in \mathrm{SE}(3),
\)
where \(v\) is the camera pose expressed in the object frame. A dataset then contains a set of such demonstration optimal views:
\(
\mathcal{V}^*_{\mathrm{demo}} = \{ v^*_{\mathrm{demo},1}, v^*_{\mathrm{demo},2}, \dots, v^*_{\mathrm{demo},N} \}
\).

\noindent \textbf{Test-time Optimal View.}  
At test time, the system selects a viewpoint from the kinematically feasible set \(\mathcal{V}\) that is \emph{closest} to the demonstration optimal view while minimizing occlusion. Formally, this is expressed as:
\begin{equation}
v^*_{\mathrm{test}} = \arg\min_{v \in \mathcal{V}} \big(d_{\mathrm{pose}}(v, v^*_{\mathrm{demo}}) + \lambda \, \mathcal{O}(\mathrm{obj}, v)\big)
\label{eq1}
\end{equation}
Here, \(\lambda\) balances the trade-off between proximity to the demonstration optimal view and occlusion, \(\mathcal{O}(\cdot)\) quantifies occlusion, \(d_{\mathrm{pose}}(\cdot,\cdot)\) measures the \(\mathrm{SE}(3)\) distance between camera poses, and \(\mathrm{obj}\) denotes the target object. Since \(v\) is expressed in the object frame, the first term encourages test-time viewpoints that are visually consistent with the training views, ensuring in-distribution observations for robust inference. Real-world examples of such optimal viewpoints are shown in Figure~\ref{optimal view images}.

\subsection{Sparse-view Gaussian Splatting}
\label{sparse gaussian}
Directly computing \(v^*_{\mathrm{test}}\) using Equation~\ref{eq1} requires depth information and an occlusion function. We show that an image-level surrogate loss provides an effective approximation. This approach entails simulating images from unseen viewpoints, which in turn requires test-time 3D reconstruction. Rather than performing time-consuming full scans, we capture sparse exploratory views of the scene, assign observer and actor roles, and construct a 3DGS representation.

\noindent \textbf{Exploratory Views.}  
We begin by capturing six scene viewpoints, evenly spaced at 60° intervals to cover the full 360° workspace. These exploratory poses are predefined and fixed. In each iteration, both arms simultaneously move to two of the poses, producing six images along with their corresponding camera poses expressed in the robot frame.

\noindent \textbf{Role Assignment.}  
Roles are assigned to each arm based on how closely their captured views match \(v^*_{\mathrm{demo}}\). Instead of performing explicit relative pose estimation, we use the robust dense feature matcher RoMa \cite{edstedt2024roma}, using the number of confident matches on the segmented object as a proxy. The arm with more aggregated matches is designated as the observer, indicating it is closer to the demonstration view, while the other arm becomes the actor.

\noindent \textbf{Three-View GS.}  
We reconstruct the scene using the three images captured by the observer, via InstantSplat \cite{fan2024instantsplat}, a sparse-view method that leverages geometric priors. It employs Mast3R \cite{leroy2024grounding}, a large-scale 3D geometric model, to estimate camera poses from RGB images and perform joint optimization. These optimized poses are more accurate than those from our low-cost robot arms, resulting in higher-quality reconstructions. We use only three images to train the 3DGS model, as the additional views provide little extra information for rendering \(v^*_{\mathrm{test}}\), and training is considerably faster. We demonstrate that three views offer the best trade-off between performance and computation time (Section~\ref{explore_view_exp}).

\noindent \textbf{Frame Alignment.}  
The 3DGS reconstruction is generated in an arbitrary coordinate system and must be aligned with the robots' frame. To achieve this, we use the Umeyama algorithm. By combining camera poses from the robot’s encoders and hand-eye calibration with the estimated poses from InstantSplat, we solve for a similarity transform via singular value decomposition (SVD) to align the two sets of poses. This process also recovers the scale of the reconstruction.

\subsection{View Optimization}
\label{view optimization}
To compute the test-time optimal view \(v^*_{\mathrm{test}}\), we first generate candidate viewpoints by global sampling within the 3DGS representation. We then select the best candidate and refine it using local gradient-based optimization. Finally, the resulting view is aligned to the real world is then aligned to the real world for the observer arm to reach.

\noindent \textbf{Candidate View Sampling.}  
To initialize the optimization, we sample a hemisphere of candidate viewpoints around the task object’s center, following \cite{breyer2022closed}. The object center is estimated from its 3D bounding box, obtained by lifting 2D masks from GroundedSAM \cite{ren2024grounded} into 3D using aligned depth maps. After filtering out kinematically infeasible poses, we obtain a candidate set \(\mathcal{V}_{\mathrm{candidate}} \subseteq \mathcal{V}\). Owing to the efficiency of 3DGS, all candidate views can be rendered at $\sim$250 FPS.

\noindent \textbf{Optimal View Initialization.}  
After rendering a set of virtual RGB images, we select the candidate view with the highest score as the initialization. The scoring function, identical to that used for role assignment, is based on the number of confidence-weighted feature matches between images from \(\mathcal{V}_{\mathrm{candidate}}\) and \(v^*_{\mathrm{demo}}\). This criterion implicitly accounts for occlusion, as occluded regions naturally produce fewer and less reliable matches. In practice, we found that this matching-based strategy outperforms an alternative based on DINOV2 \cite{oquab2024dinov2} cosine distance for measuring view similarity. Even when combined with object segmentation, DINOV2 struggles to discriminate between viewpoints—an observation consistent with \cite{NEURIPS2024_23866f14}. To reduce computational cost, we employ Tiny RoMa \cite{edstedt2024roma}, a lightweight variant of RoMa with faster runtime.

\begin{figure*}[t!]
\centering
\includegraphics[width=1.0\textwidth]{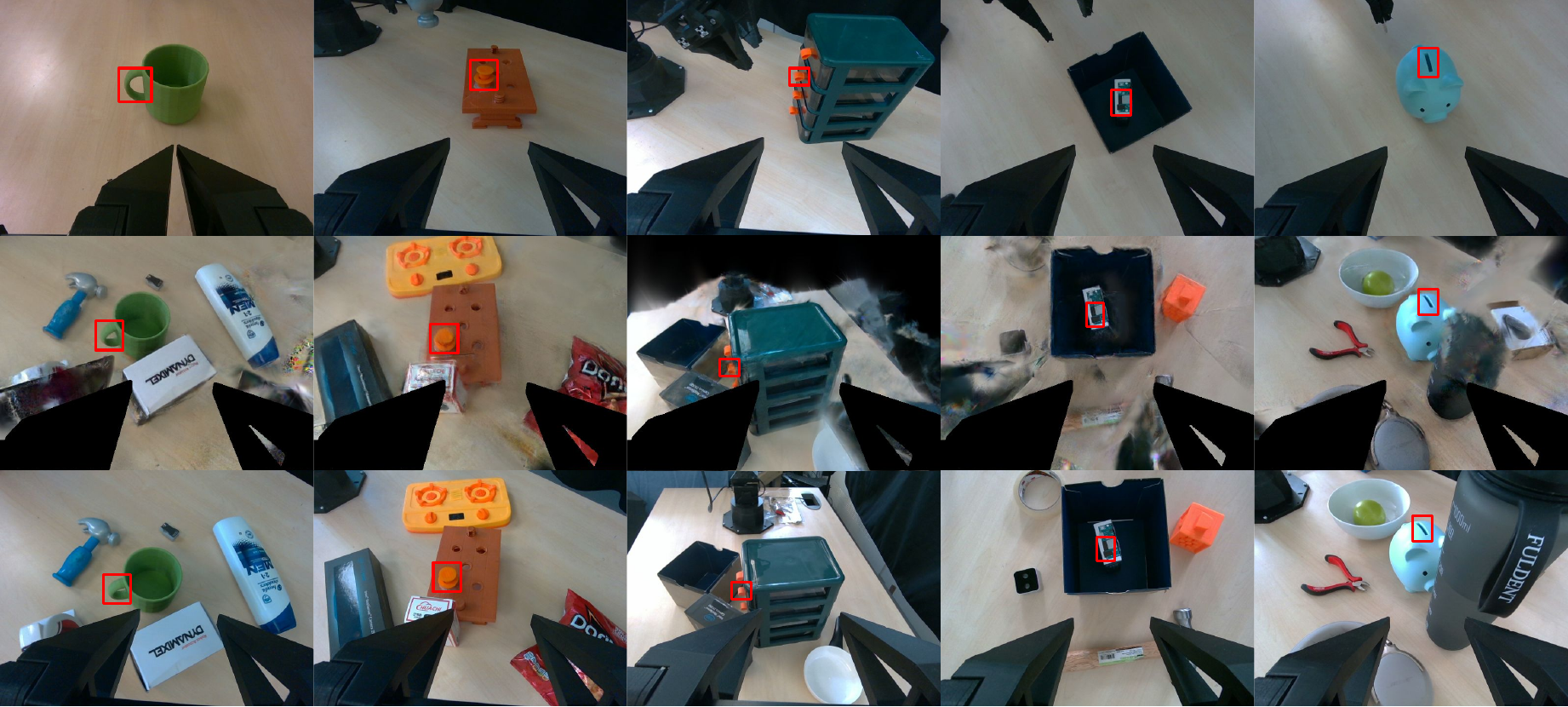}
\caption{\textbf{Images of Optimal Views.} Top row: demonstration optimal views. Middle row: test-time optimal views in 3DGS with gripper mask overlay. Bottom row: real world test-time optimal views. Red boxes indicate the task-relevant object parts. Test-time optimal views are derived by reconstructing the demonstration’s optimal viewpoints subject to minimal occlusion.}

\label{optimal view images}
\vspace{-10pt}
\end{figure*}

\noindent \textbf{Differentiable Rendering.}  
After obtaining an initialization, we refine the viewpoint \(v\) using differentiable rendering while explicitly accounting for gripper-induced occlusions. This is particularly important in our setting, where the observer arm’s gripper consistently remains in the field of view and may block the object, as shown in Figure~\ref{optimal view images}. To handle this, we use SAM2 \cite{ravisam} for segmentation of the rendered image, guided by the demonstration image's mask as a prompt. This is effective because the initialized and demonstration viewpoints are already closely aligned. To mitigate errors in estimating the object center, we first re-center the object’s mask in the image plane. We then refine \(v\) by optimizing a loss that aligns the image rendered from the 3DGS model (denoted \(\mathcal{G}\)) with the segmented demonstration image \(I^*_{\mathrm{demo}}\), while penalizing occlusions. Let \(I(v; \mathcal{G})\) denotes the rendered image. The loss $\mathcal{L}$ as a particular view $v$ is defined as:
\begin{equation}
\begin{aligned}
\mathcal{L}(v) &= -\lambda_1 \, \mathcal{L}_{\mathrm{sim}}\big(\phi(I(v; \mathcal{G}) \odot \mathcal{M}_{\mathrm{obj}}(v)\big), \phi(I^*_{\mathrm{demo}})\big) \\
&\quad + \lambda_2 \, \mathcal{L}_{\mathrm{IOU}}(\mathcal{M}_{\mathrm{obj}}(v), \mathcal{M}_{\mathrm{grip}}),
\end{aligned}
\label{diff_render_eq}
\end{equation}
where \({\phi}(\cdot)\) denotes the DINOV2 feature extractor, \(\mathcal{M}_{\mathrm{obj}}(v)\) is the soft object mask at viewpoint \(v\) obtained from SAM2, and \(\mathcal{M}_{\mathrm{grip}}\) is the fixed soft gripper mask. The first term enforces feature alignment, while the second penalizes overlap between the object and gripper masks. Unlike photometric losses employed in prior work~\cite{yen2021inerf}, our loss explicitly accounts for occlusions. We note although DINOV2 is unreliable for global initialization, it is effective for local refinement.

\noindent \textbf{Moving to the Optimal View.}  
Once the optimal camera pose \(v^*_{\mathrm{test}}\) is computed, we transform it back into the real-world coordinate frame and use a motion planner to move the observer arm to the target pose. Figure~\ref{optimal view images} illustrates the demonstration view, the optimal view obtained from the 3DGS representation, and the final aligned view executed in the real world. Minor discrepancies in viewpoint arise from robot kinematic errors, hand–eye calibration, and frame alignment, but these do not noticeably affect downstream imitation learning performance.

\subsection{View-Conditioned Imitation Learning}
\label{imitation learning}
In our framework, we extend two categories of imitation methods for task execution once the camera has been moved to the optimal view: trajectory transfer and behavior cloning.

\noindent \textbf{AV Trajectory Transfer.}  
Trajectory transfer (TT) methods estimate the relative pose change of the object between the demonstration and the test instance, and then transfer the demonstration trajectory in a one-shot manner within the $\mathrm{SE}(3)$ manifold \cite{doi:10.1126/scirobotics.adv7594}. Consequently, only a single demonstration optimal view is required. In our setting, demonstrations are recorded in the actor arm’s coordinate frame, while observations are captured from the observer arm’s camera. To resolve this mismatch, we extend trajectory transfer by applying the following sequence of frame transformations:
\begin{align}
    ^{\mathrm{O}}\mathbf{T}_{\mathrm{E_O}}(\mathrm{test}) &= 
    {}^{\mathrm{O}}\mathbf{T}_{\mathrm{E_O}} \, 
    {}^{\mathrm{E_O}}\mathbf{T}_{\mathrm{C}} \, \Delta{}^{\mathrm{C}}\mathbf{T}_{\mathrm{obj}} \, {}^{\mathrm{C}}\mathbf{T}_{\mathrm{E_O}} \\
    \Delta{}^{\mathrm{O}}\mathbf{T}_{\mathrm{obj}} &=  {}^{\mathrm{O}}\mathbf{T}_{\mathrm{E_O}}(\mathrm{test}) \,
    ^{\mathrm{E_O}}\mathbf{T}_{\mathrm{O}}(\mathrm{demo}) \\
    \Delta{}^{\mathrm{A}}\mathbf{T}_{\mathrm{obj}} &= {}^{\mathrm{O}}\mathbf{T}_{\mathrm{A}} \, \Delta{}^{\mathrm{O}}\mathbf{T}_{\mathrm{obj}} \, {}^{\mathrm{A}}\mathbf{T}_{\mathrm{O}} \\
    {}^{\mathrm{A}}\mathbf{T}_{\mathrm{E_A}}(\mathrm{test}) &= \Delta{}^{\mathrm{A}}\mathbf{T}_{\mathrm{obj}} \, {}^{\mathrm{A}}\mathbf{T}_{\mathrm{E_A}}(\mathrm{demo})
\end{align}
Here, ${}^{\mathrm{O}}\mathbf{T}_{\mathrm{E_O}}(\mathrm{test})$ is the observer end-effector pose at $v^*_{\mathrm{test}}$, and ${}^{\mathrm{O}}\mathbf{T}_{\mathrm{E_O}}$ is its pose at $v^*_{\mathrm{train}}$, respectively, both expressed in the observer base frame $\mathrm{O}$. The hand–eye calibration from the observer end-effector to the camera frame is ${}^{\mathrm{E_O}}\mathbf{T}_{\mathrm{C}}$, with inverse ${}^{\mathrm{C}}\mathbf{T}_{\mathrm{E_O}}$. The object’s relative pose change from demonstration to test is denoted by $\Delta{}^{\mathrm{C}}\mathbf{T}_{\mathrm{obj}}$ in the camera frame, $\Delta{}^{\mathrm{O}}\mathbf{T}_{\mathrm{obj}}$ in the observer base frame, and $\Delta{}^{\mathrm{A}}\mathbf{T}_{\mathrm{obj}}$ in the actor base frame $\mathrm{A}$. The fixed transform between the observer and actor base frames is ${}^{\mathrm{O}}\mathbf{T}_{\mathrm{A}}$, with inverse ${}^{\mathrm{A}}\mathbf{T}_{\mathrm{O}}$. Finally, ${}^{\mathrm{A}}\mathbf{T}_{\mathrm{E_A}}(\mathrm{demo})$ and ${}^{\mathrm{A}}\mathbf{T}_{\mathrm{E_A}}(\mathrm{test})$ denote the actor’s end-effector poses for task execution during demonstration and test, both expressed in the actor base frame $\mathrm{A}$.

We estimate $\Delta{}^{\mathrm{C}}\mathbf{T}_{\mathrm{obj}}$ using RoMa by lifting matched feature points to 3D via aligned depth maps. The relative $\mathrm{SE}(3)$ transformation is then computed using Procrustes alignment within a RANSAC loop. The transferred trajectory is executed in open-loop, and its accuracy further depends on the hand–eye calibration ${}^{\mathrm{E_O}}\mathbf{T}_{\mathrm{C}}$ and the fixed transform between the observer and actor base frames ${}^{\mathrm{O}}\mathbf{T}_{\mathrm{A}}$.

\noindent \textbf{AV Behavior Cloning.}  
Behavior cloning (BC) trains a closed-loop policy $\pi$ to map observations to actions. Unlike TT, which exploits pose estimation models, BC requires several demonstration optimal views to generalize across diverse camera perspectives covering $v^*_{\mathrm{test}}$. Since optimizing over all viewpoints in $\mathcal{V}^*_{\mathrm{train}}$ is computationally expensive, we use Mast3R to estimate camera poses in the object frame and select the viewpoint closest to the center of all estimated poses as $v^*_{\mathrm{demo}}$, which we found to work well empirically. Moreover, unlike prior work~\cite{pmlr-v270-cheng25b, chuang2024active}, which represents the end-effector pose in a static world frame and provides the camera pose explicitly as policy input, we express the actor arm’s end-effector pose directly in the camera frame:
\begin{equation}
    {}^{\mathrm{C}}\mathbf{T}_{\mathrm{E_A}} =
    {}^{\mathrm{C}}\mathbf{T}_{\mathrm{E_O}} \,
    {}^{\mathrm{E_O}}\mathbf{T}_{\mathrm{O}} \,
    {}^{\mathrm{O}}\mathbf{T}_{\mathrm{A}} \,
    {}^{\mathrm{A}}\mathbf{T}_{\mathrm{E_A}} .
\end{equation}
This representation simplifies the state space and, as we show in Section~\ref{action_in_cam_exp}, improves both data efficiency and policy performance.  
At each time step \(t\), the policy receives an RGB observation \(I_t\) from the test-time optimal view \(v^*_{\mathrm{test}}\), along with the proprioceptive state \(S_t\).  
The proprioceptive state includes the actor end-effector pose expressed in the camera frame and the gripper state.  
The policy outputs an action sequence
\(
a_{t:t+n_p-1} = \pi \big( I_t,\, S_t \big)
\)
of length \(n_p\), which is then transformed back to the actor’s base frame for execution.

\noindent \textbf{Ambidextrous Inference.}
By representing actions in the camera frame for both of our imitation learning methods—AV TT and AV BC—we enable ambidextrous inference: when the roles of observer and actor differ from those in the demonstration, the transferred trajectory and trained policy can still be executed without any additional data or training.

\begin{table*}[ht!]
\centering
\caption{Success Rates of Different Methods on the Five Tasks (without and with occlusions).}
\begin{tabular}{lcccccccccccc}
\toprule
\textbf{Method} & \multicolumn{2}{c}{\textbf{Pick Mug}} & \multicolumn{2}{c}{\textbf{Hammer Nail}} & \multicolumn{2}{c}{\textbf{Open Drawer}} & \multicolumn{2}{c}{\textbf{Retrieve Pack}} & \multicolumn{2}{c}{\textbf{Insert Coin}} & \multicolumn{2}{c}{\textbf{Sum}} \\
 & no occ & occ & no occ & occ & no occ & occ & no occ & occ & no occ & occ  & no occ & occ \\ 
\midrule
TT (3-views)  &  5/10  &  3/10  &   4/10   &   4/10   &  2/10  &  1/10    &  N/A   &   1/10    &  0/10  &  0/10  &   11/40   &  9/50   \\
TT (ours with AV)  &   \textbf{8/10}    &  \textbf{8/10}  &   \textbf{10/10}   &  \textbf{7/10}  &  \textbf{6/10}   &  \textbf{4/10}   &   N/A   &  \textbf{9/10} &      \textbf{3/10} &   \textbf{2/10}   &   \textbf{27/40}    &  \textbf{30/50}    \\
\midrule
BC (static camera)   &    4/10  &  4/10    &  3/10   &   2/10   &  1/10 &       1/10 &  N/A  &  0/10 &  0/10    &   0/10   &   8/40   &  7/50     \\
BC (ours with AV)   &   \textbf{6/10}   &  \textbf{5/10}    &   \textbf{4/10}   &   \textbf{5/10}   &  \textbf{3/10} &  \textbf{3/10}  &   N/A  &  \textbf{3/10}  & \textbf{1/10}  &   \textbf{1/10}   &  \textbf{14/40}   &  \textbf{17/50}   \\
\bottomrule
\end{tabular}
\label{tab:manipulation_success_occlusion}
\end{table*}

\section{Evaluation}
We instantiated our method on a real-world dual-arm ALOHA setup~\cite{zhao2023learning}, mounted on a table and equipped with two calibrated RealSense D405 cameras. To evaluate our approach, we selected five diverse manipulation tasks involving self-occluding objects. In \emph{Pick Mug}, the robot grasps a mug by its handle. In \emph{Hammer Nail}, it uses a pre-grasped hammer to drive a nail. \emph{Open Drawer} requires opening the second drawer of a multi-layer unit. In \emph{Retrieve Pack}, the robot retrieves a package from a deep box. Finally, \emph{Insert Coin} involves placing a coin into a storage container. All tasks require observing specific object parts or features for successful manipulation, with the key regions highlighted in red in Figure~\ref{optimal view images}.

\subsection{Comparison with Static-Camera}

\begin{figure*}[t!]
\centering
\includegraphics[width=1.\textwidth]{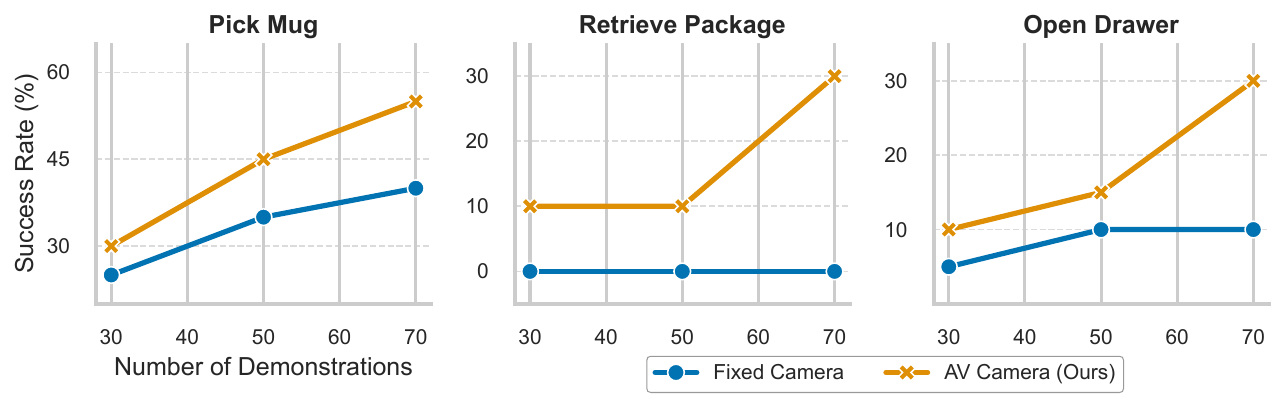}
\caption{\textbf{Data Efficiency of Behavior Cloning with Active Vision.} With the same number of demonstrations, our method outperforms the static camera setup across three tasks, evaluated using 30, 50, and 70 demonstrations.}
\label{fig:scale}
\vspace{-10pt}
\end{figure*}

\noindent \textbf{Baselines and Implementation.}  
We evaluate our method for both TT and BC approaches, comparing against static-camera setups. For the TT baseline, we select the best image from the initial 3-view exploration and perform pose estimation relative to the demonstration, as described in Section~\ref{imitation learning}. In contrast, our method leverages the test-time optimal view for pose estimation. For the BC baseline, we use a static camera chosen from one of the poses in the 3-view exploration.  

To ensure fairness, we collect separate datasets for BC with and without active vision, rather than reusing the same dataset as was done in \cite{chuang2024active}. Reusing would introduce visual inconsistencies, as the static camera would capture the observer arm moving across different poses, providing irrelevant information that could confuse the learned policy. In total, we collected 70 demonstrations for each of the five tasks, with and without active vision, resulting in \(70 \times 5 \times 2 = 700\) demonstrations. All demonstrations were recorded using an iPhone-based teleoperation system built with ARKit.

We implement BC with ACT \cite{zhao2023learning}, using DINOv2 as the vision backbone and absolute Cartesian action representation. For BC with AV, we remove any frames in the dataset where the actor arm’s gripper is invisible in the camera view, as these introduce ambiguities. This also ensures that the trained policy outputs a pose where the actor’s gripper is likely to be visible in the camera view at the first timestep.

\noindent \textbf{Experiments.}  
For all tasks except \emph{Retrieve Pack}, which inherently involves occlusion due to its setup, we evaluate both unoccluded and occluded scenarios. We perform 10 rollouts on novel scene configurations with variations in object poses and the presence of distractors. For each variation, we first run the baseline and then our method under identical conditions.  

For the static-camera BC baseline, both training and evaluation are restricted to object poses where task-relevant parts remain visible, as self-occlusion leads to prohibitively low success rates. Consequently, pose variations for the static-camera BC baseline are limited to within 45° of the demonstration poses during testing. In contrast, our method naturally handles self-occlusion arising from larger rotations.

\noindent \textbf{Results.} Table~\ref{tab:manipulation_success_occlusion} summarizes the experiment results. Using the test-time optimal view for inference, both TT and BC consistently outperform the static-camera setup across all tasks. Specifically, TT achieves performance improvements of 146\% in the unoccluded setting and 233\% in the occluded setting, while BC improves by 75\% and 143\%, respectively.

For TT, this improvement arises because the test-time optimal image has greater feature overlap with the demonstration optimal image and experiences less occlusion, leading to more accurate pose estimation. For BC, the test-time observations are more in-distribution relative to the training data and encounter fewer occlusions, resulting in more reliable policy execution. The performance drop under occlusion may stem from selected viewpoints being slightly out-of-distribution in order to compensate for occluded regions.

\noindent \textbf{Data Efficiency of BC with AV.} Figure~\ref{fig:scale} shows the success rates versus the number of demonstrations, using 30, 50, and 70 demos for the \emph{Pick Mug}, \emph{Retrieve Pack}, and \emph{Open Drawer} tasks. We observe that, given the same number of demonstrations, BC with AV consistently outperforms the static-camera setup, highlighting the advantages of inference from test-time optimal views. Notably, for the \emph{Retrieve Pack} task, BC with a static camera fails completely. This is because the task suffers from severe occlusion, particularly from the actor arm’s gripper when reaching inside the box. Some training frames contain the package fully occluded by the gripper, which confuses the policy and prevents it from learning reliable grasp decisions.


\noindent \textbf{Failure Modes.} Our method can fail if the AV pipeline selects a suboptimal view. In practice, the pipeline typically generates views that are considered reasonable by humans, validating the effectiveness of sparse-view 3DGS for active vision. However, we occasionally observe occlusion of critical regions, such as drawer handles during rollouts, because the current formulation does not explicitly enforce full visibility of key object parts. Downstream imitation learning methods can also fail despite receiving a good view. TT with AV may fail due to inaccurate feature matching or calibration errors—either from hand–eye calibration or from the relative calibration between the observer and actor arms. For BC with AV, failures mostly occur when the actor arm's initial end-effector state in the camera frame deviates from those seen in the demonstrations, causing compounding errors. This is more likely when the test-time optimal views differ significantly from the demonstration view. Conversely, tasks typically succeed when the actor’s gripper is sufficiently close to the target object, as fine-grained actor–object interactions are accurately captured in the optimal demonstration-view dataset. Finally, because the learned policy relies on a single RGB observation, it remains sensitive to depth ambiguities.







Experiment videos are available on our project webpage at \hypersetup{urlcolor=blue} \url{https://obact.github.io}, including demonstrations of dynamic role assignment, timed task executions, object generalization, and object tracking capabilities of our system.

\subsection{Representing Actions in the Camera Frame}
\label{action_in_cam_exp}
We evaluate the effect of representing actions in the camera frame for BC with AV. We compare against prior approaches~\cite{chuang2024active, xiong2025vision}, where actions are represented in the fixed actor robot frame and the camera pose is provided as an input. Results in Table~\ref{tab:camera_frame_ablation} show that representing actions directly in the camera frame improves both generalization and task success. We hypothesize that providing the camera pose as input is less effective because object poses span a wide distribution while camera views remain relatively similar, making it difficult for the policy to infer the robot’s pose in the static frame. On the other hand, representing actions directly in the camera frame provides a consistent reference.

\begin{table}[h]
\centering
\caption{Impact of action representation on success rates.}
\label{tab:camera_frame_ablation}
\begin{tabular}{lcc}
\toprule
\textbf{Method} & \textbf{Mug} & \textbf{Hammer}\\
\midrule
Camera pose as input, action in robot frame & 1/10 & 0/10 \\
Action in camera frame (ours) & \textbf{6/10} & \textbf{4/10} \\
\bottomrule
\end{tabular}
\end{table}

However, representing actions in the camera frame also introduces a potential issue. In practice, we found that the training viewpoints must be sufficiently varied; otherwise, the model can shortcut learning by relying primarily on the camera pose. When trajectories relative to the camera are too similar, the policy may ignore the image inputs entirely.

\subsection{Impact of Number of Exploration Views}
\label{explore_view_exp}
\begin{figure}[t!]
\centering
\includegraphics[width=.47\textwidth]{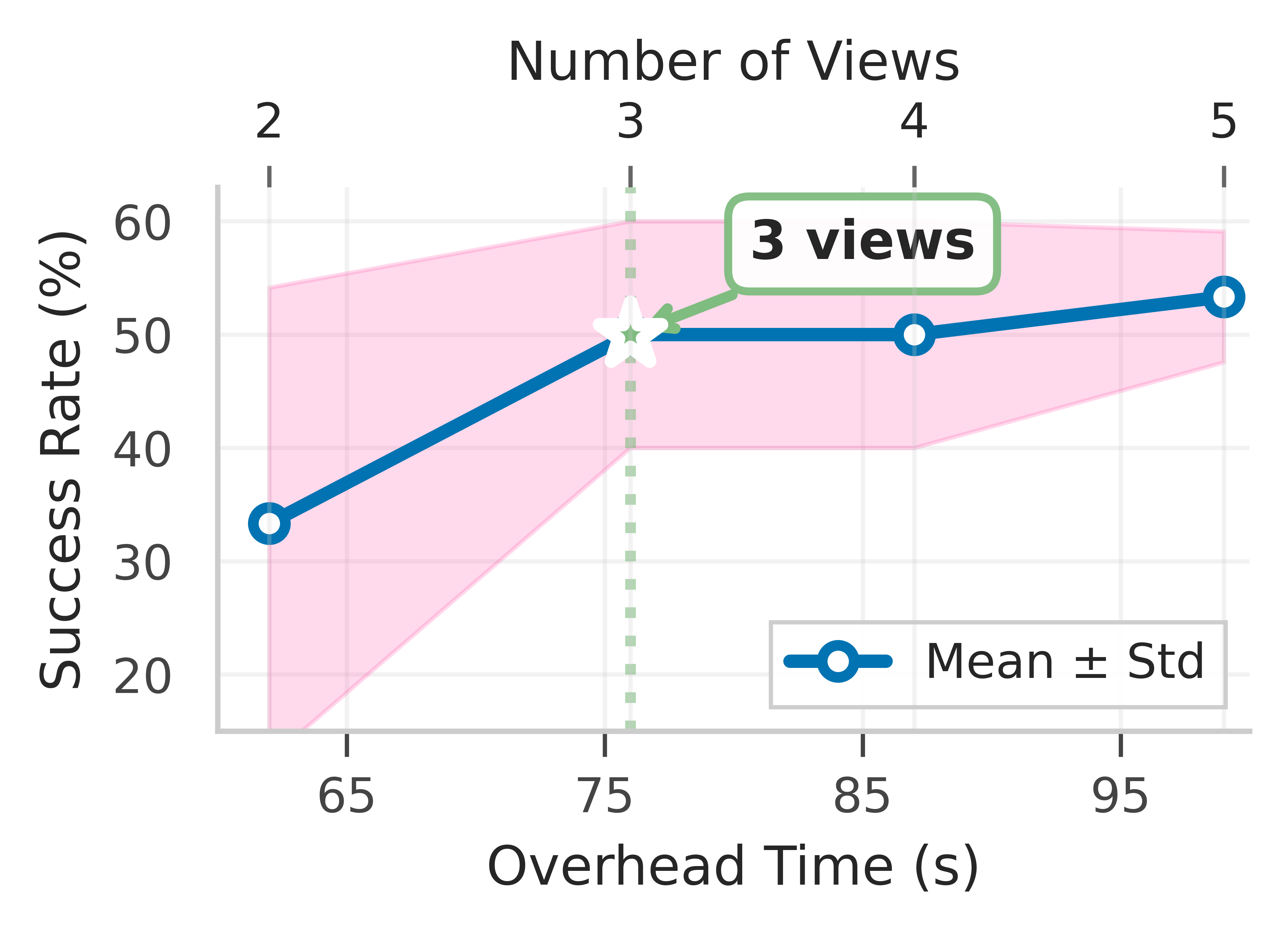}
\caption{Effect of Number of Exploration Views.}
\label{fig:number_of_views}
\vspace{-10pt}
\end{figure}
We evaluate the impact of exploring three views per arm at test time to acquire scene information. Specifically, we compare 2–5 exploration viewpoints per arm to study how the extent of visual coverage affects downstream manipulation success. We use BC with AV and evaluate on the \emph{Pick Mug} task under occlusion, with results shown in Figure~\ref{fig:number_of_views}. Our findings indicate that three views strike a good balance between performance and execution time, providing sufficient scene coverage for reliable role assignment and optimal viewpoint selection. Since frame alignment requires at least three views, using only two necessitates relying on camera poses inferred from the robot’s encoders and hand–eye calibration, bypassing full alignment. This leads to higher variance in performance because the resulting 3DGS reconstructions are less accurate, widening the sim-to-real gap. Consequently, test-time optimal views may also be less reliable. This limitation could be mitigated with more precise robots and better calibration.

\begin{table}[h]
\centering
\caption{Time breakdown for each component of our AV pipeline, measured with a RTX 4080Ti GPU.}
\begin{tabular}{l c}
\toprule
\textbf{Component} & \textbf{Time (s)} \\
\midrule
Six-view Exploration              & 18 \\
InstantSplat Geometric Initialization & 20 \\
InstantSplat 3DGS Training      & 23 \\
Optimal View Initialization             & 3  \\
Differential Rendering          & 12 \\
\midrule
\textbf{Total}                  & 76 \\
\bottomrule
\end{tabular}
\label{tab:time_breakdown}
\end{table}

Table~\ref{tab:time_breakdown} reports the time breakdown of our AV pipeline. The majority of time is spent on InstantSplat, while the exploration step could be further reduced by increasing the arms' speed. We believe that as sparse-view 3DGS methods continue to improve, our system will become more efficient.




\section{Conclusions}
We have present ObAct, a novel observer–actor framework for imitation learning in active vision, in which an observer arm computes and moves to optimal visual observations to guide the actor arm’s actions. Our method captures fine-grained gripper–object interactions, supporting robust manipulation across diverse scenarios using both trajectory transfer and behavior cloning. Experimental results demonstrate significant gains in success rates compared to a static camera setup across both unoccluded and occluded scenarios.

Despite the effectiveness of our method, several limitations remain. The active-vision pipeline is relatively slow, the approach is tailored to short-horizon tasks, and it lacks reactivity to environmental changes. Moreover, the current setup cannot handle tasks that require two arms acting simultaneously. We identify several promising directions for future work. One is to explore dynamic viewpoints that continuously track gripper–object interactions during execution, enabling richer visual feedback and closed-loop occlusion avoidance. Another is to extend our approach to long-horizon tasks and to deformable-object manipulation, both of which introduce additional challenges for active vision and imitation learning. Finally, for dual-arm manipulation, we envision expanding the system to a three-arm configuration, in which one arm dynamically acts as the observer while the remaining two serve as the manipulators. We believe these extensions will enable more resilient active-vision robotic systems capable of handling complex manipulation tasks in diverse, unstructured real-world environments.


{\
\printbibliography
}

\end{document}